\documentclass[sn-mathphys-num, iicol]{sn-jnl}

\usepackage{graphicx}%
\usepackage{multirow}%
\usepackage{amsmath,amssymb,amsfonts}%
\usepackage{amsthm}%
\usepackage{mathrsfs}%
\usepackage{xfrac}
\usepackage{arydshln}
\usepackage{hyperref}
\usepackage{xurl}
\usepackage[title]{appendix}%
\usepackage[dvipsnames]{xcolor}%
\usepackage{textcomp}%
\usepackage{manyfoot}%
\usepackage{booktabs}%
\usepackage{float}
\usepackage{algorithm}%
\usepackage{algorithmicx}%
\usepackage{algpseudocode}%
\usepackage{listings}%
\usepackage{soul}%

\theoremstyle{thmstyleone}%
\theoremstyle{thmstyletwo}%

\theoremstyle{thmstylethree}%

\begin{document}

\title[Investigation of the Impact of Synthetic Training Data in the Industrial Application of Terminal Strip Object Detection]{Investigation of the Impact of Synthetic Training Data in the Industrial Application of Terminal Strip Object Detection}


\author*[1]{\fnm{Nico} \sur{Baumgart}}\email{nico.baumgart@th-owl.de}

\author[1]{\fnm{Markus} \sur{Lange-Hegermann}}\email{markus.lange-hegermann@th-owl.de}
\equalcont{These authors contributed equally to this work.}

\author[2]{\fnm{Mike} \sur{Mücke}}\email{mmuecke@phoenixcontact.com}
\equalcont{These authors contributed equally to this work.}

\affil[1]{\orgdiv{Department of Electrical Engineering and Computer Science}, \orgname{OWL University of Applied Sciences and Arts}, \orgaddress{\street{Campusallee 12}, \city{Lemgo}, \postcode{32657}, \state{North Rhine-Westphalia}, \country{Germany}}}

\affil[2]{\orgdiv{Chief Digital Office - Digital Innovations}, \orgname{Phoenix Contact GmbH \& Co. KG}, \orgaddress{\street{Flachsmarktstraße 8}, \city{Blomberg}, \postcode{32825}, \state{North Rhine-Westphalia}, \country{Germany}}}

\abstract{In industrial manufacturing, deploying deep learning models for visual inspection is mostly hindered by the high and often intractable cost of collecting and annotating large-scale training datasets. While image synthesis from 3D CAD models is a common solution, the individual techniques of domain and rendering randomization to create rich synthetic training datasets have been well studied mainly in simple domains. Hence, their effectiveness on complex industrial tasks with densely arranged and similar objects remains unclear. In this paper, we investigate the sim-to-real generalization performance of standard object detectors on the complex industrial application of terminal strip object detection, carefully combining randomization and domain knowledge. We describe step-by-step the creation of our image synthesis pipeline that achieves high realism with minimal implementation effort and explain how this approach could be transferred to other industrial settings. Moreover, we created a dataset comprising 30.000 synthetic images and 300 manually annotated real images of terminal strips, which is publicly available for reference and future research. To provide a baseline as a lower bound of the expectable performance in these challenging industrial parts detection tasks, we show the sim-to-real generalization performance of standard object detectors on our dataset based on a fully synthetic training. While all considered models behave similarly, the transformer-based DINO model achieves the best score with 98.40\% mean average precision on the real test set, demonstrating that our pipeline enables high quality detections in complex industrial environments from existing CAD data and with a manageable image synthesis effort.}




\maketitle

\section{Introduction}
\label{sec:introduction}
Currently, various research fields in machine learning~(ML) are evolving rapidly, setting new benchmarks and demonstrating the potential benefit of this technology for society. In particular, the availability of computing power and large-scale public datasets advanced the development and initiated the era of deep learning (DL). However, the amount of data required to train DL models limits the transferability of current state-of-the-art approaches to real-world applications in which data is scarce and difficult to gather~\cite{Tsirikoglou.2020}, such as computer vision (CV) applications in industrial manufacturing. 

The main challenge arises from the manual effort to collect and label the training images for supervised learning. For instance, the creators of Microsoft Common Objects in Context (MS COCO), a commonly used benchmark dataset for object detection and semantic segmentation, stated in~\cite{Lin.2014} that it took around 60,000 worker hours to gather and annotate these images, which is intractable for most industrial applications. Moreover, industrial settings are dynamic, so the environment and the objects of interest may change, forcing the training data to be dynamically adaptable within a reasonable time. Therefore, recent works focused on the usage of simulated images for model training and investigated their impact on model performance in real-world test cases~\cite{Akar.2022,Ljungqvist.2023,Zhu.2023}.

For industrial manufacturers, so-called synthetic training data offers the following advantages:

\begin{itemize}
    \item Accurate 3D models for image synthesis are often available from computer-aided design (CAD), used during the construction of products.
    \item Theoretically, an arbitrary amount of training images can be generated using an automated image synthesis pipeline.
    \item Ground-truth annotations can be calculated automatically, eliminating the human source of error and saving costs for manual annotation.
    \item The training data can be generated without real-world biases by adapting parameters in the image synthesis pipeline.
\end{itemize}
\noindent
However, the domain gap between synthetic and real images often causes the DL models to perform worse in real-world test cases, so numerous approaches have been proposed to close this gap.

Apart from the usual data augmentation techniques and the related cut-and-paste approach~\cite{Dvornik.2018}, domain adaptation and domain randomization (DR) are the two main concepts currently used in this area based on rendered images of 3D models. The former maps one domain into the other or both into an intermediate third domain, using, e.g., GANs~\cite{Rojtberg.2020,Shrivastava.2017,Wang.2022}. Thus, domain adaptation still relies on real images, which might be unsuitable for industrial use cases. In contrast, DR is based on random, even unrealistic variations, e.g., of textures, backgrounds, or lighting conditions, to suggest to the DL model that real images are just another variation of the learned domain~\cite{Tobin.2017}. Hence, it aims to generalize directly to the real world and enables fully synthetic training, which has already shown promising results in different use cases, such as detecting cars~\cite{Tremblay.2018}, small load carriers~\cite{Mayershofer.2021}, and turbine blades at a manual working station~\cite{Eversberg.2021}. Nevertheless, experiments in industrial environments were conducted under highly simplified conditions, including a small amount of well-separable classes and considering only similar-sized objects. The expectable performance of state-of-the-art approaches on more complex industrial tasks in which the objects of interest could be very similar and densely arranged remains unclear. 

In this paper, we assess the particularly challenging object detection of terminal strips using 3D models by Phoenix Contact that are publicly available on the company's homepage\footnote{\url{https://www.phoenixcontact.com/en-pc/products/terminal-blocks/feed-through-terminal-blocks-multi-conductor-terminal-blocks-and-multi-level-terminal-blocks}}. The investigation is motivated by two applications: identification of already mounted terminal blocks in electrical cabinets and quality control in terminal strip assemblies. The first application is relevant, as the product information is printed on the side of the terminal blocks and occluded by neighboring components in a mounted state. Thus, experts must identify certain terminal blocks to avoid demounting. The second application of quality control stems from the fact that the product portfolio for terminal strips is manifold, including over 10,000 components. This causes manual work steps during terminal strip assembly. Since manual processes are prone to human error and experts are not always available, a ML-based detection system would prevent errors and save time in both cases.

We approached this task by building an automated image synthesis pipeline and generating 30,000 synthetic images of random terminal strips. Thereby, we focused on including domain specific characteristics, e.g., the arrangement of the objects of interest relative to each other and their colors, while randomizing the lighting conditions, viewpoints, and background to keep the modeling effort manageable. This procedure could also be transferred to other industrial applications, as we discuss in Section~\ref{sec:discussion}. Moreover, we annotated 300 handmade real images for evaluation and provide baseline results of four standard object detectors, without excessive hyperparameter tuning and entirely trained on synthetic images. This serves as a lower bound of the expectable sim-to-real generalization performance since any labeled ground truth data of the target domain would presumably lead to an improvement.

As we found that scaling is most crucial for the sim-to-real domain gap in this specific use case, we also show how synthetic data can be used to learn a dynamic image preprocessing that significantly improves the real-world detection performance. To motivate further developments on this use case, we made our dataset publicly available on Zenodo\footnote{\url{https://zenodo.org/records/10674995}} for research purposes. 

The remainder of this paper is structured as follows. Section~\ref{sec:relwork} points out usual image synthesis methods and available synthetic image datasets to show recent developments in sim-to-real DL. Then, we describe in Section~\ref{sec:pipeline} how DR and domain knowledge are combined in our pipeline to create a synthetic image dataset for terminal strip object detection. While we detail our experiments to assess the sim-to-real domain gap and present the corresponding results in Section~\ref{sec:evaluation}, Section~\ref{sec:discussion} discusses the significance of the results for industrial manufacturers, the transferability of our strategy to other industrial scenarios, and how synthetic data performs against augmented data in the worst case consideration of no labeled ground truth images of the target domain being available. The paper is concluded in Section~\ref{sec:conclusions} also giving a brief outlook for future work.

\section{Related Work}
\label{sec:relwork}
This section briefly discusses popular synthetic image datasets and recent image synthesis methods to cover the current state-of-the-art of using simulated data for visual DL.

\subsection{Synthetic Image Datasets}
Although annotated synthetic image datasets can be generated much faster than their real-world counterparts, they are still outnumbered and only available for specific use cases. One very active research area dealing with simulated training data is autonomous driving, in which different approaches are used for dataset creation. In~\cite{Richter.2016}, e.g., the authors exploited the video game engine of GTA5 to obtain 25,000 photorealistic street scene images with pixel-level segmentation masks. By substituting \sfrac{2}{3} of the real training images, they could improve the performance of a semantic segmentation model, which demonstrates the potential benefit of synthetic data.

Contrarily, the creators of Virtual KITTI semi-automatically reconstructed real-world street scenes from the original KITTI dataset~\cite{Geiger.2012} and varied several parameters, such as the weather conditions, resulting in 35 synthetic videos with approximately 17,000 frames labeled for various video understanding tasks~\cite{Gaidon.2016}. Their results show that a combination of synthetic pre-training and real fine-tuning yields better performance than each separate approach in multi-object tracking, supporting the importance of simulated data. 

Another notable street scene dataset is Synscapes~\cite{Wrenninge.2018} which represents the synthetic counterpart to Cityscapes~\cite{Cordts.2016}. In contrast to the previously mentioned datasets, Synscapes was created using a parameter-based random generator to build a new scene for each image. Furthermore, the developers leveraged photorealistic rendering techniques and found that high image realism is necessary to bridge the sim-to-real domain gap~\cite{Wrenninge.2018}. Nevertheless, the application of autonomous driving is out of scope for industrial manufacturers, so the observed results may not apply to their use cases.

The amount of synthetic image datasets for industrial requirements is small, even though several approaches for image synthesis have been proposed and evaluated in small-scale industrial test cases, as discussed in the following subsection. Two recently published datasets are SIP-17~\cite{Zhu.2023} and SORDI~\cite{Akar.2022}, aiming to close this gap. SIP-17 is a synthetic dataset for industrial parts classification consisting of 17 objects in six different use cases to allow research in close-to-reality scenarios. The experiments reveal that classification is difficult when classes share a similar albedo or have a comparable shape~\cite{Zhu.2023}, which is typical for industrial objects.

While SIP-17 contains 33,000 images, SORDI was presented as the first large-scale synthetic image dataset for the industry with 200,000 images at the publication date. It is characterized by eight industrial assets in 32 fully synthetic factory scenarios and enables the sim-to-real benchmarking of vision models for intelligent robots in factory environments~\cite{Akar.2022}. However, SORDI cannot serve the entire range of industrial applications, so further work is required to cover, e.g., quality control in assembly and object tracking.

To conclude, it is difficult for industrial manufacturers to estimate how a synthetically trained DL model would perform in a specific use case, as there is not enough data available to benchmark sim-to-real methods in different industry-related applications. Thus, we provide our dataset to enable further investigations in this area with a focus on narrow and very similar industrial objects that are densely arranged and partly occlude each other.

\subsection{Image Synthesis for Deep Learning}
Generating synthetic training images for supervised learning tasks requires suitable design choices to improve the sim-to-real generalization performance. These choices include textures of the objects of interest, the image background, different viewpoints, lighting conditions, etc., potentially making the modeling of virtual scenes very time-consuming and complex as well. Especially when the reality should be closely approximated, modeling the entire range of possible real-world variations might eliminate the advantage of saving time for data annotation. Hence, there are various approaches to apply image synthesis that attempt to minimize the domain gap while avoiding excessive modeling effort. For instance, the authors of~\cite{Dwibedi.2017} used segmentation masks of real images to cut out the objects of interest and place them onto random background images, which can be seen as image synthesis. This idea was later improved in~\cite{Dvornik.2018} proposing a context model that provides patches of the background images likely to contain certain objects to create more realistic training data. Although this method is very efficient and showed promising results in object detection~\cite{Dvornik.2018}, it relies on real-world data which might not be available or too expensive to gather in industrial applications. Alternatively, the appearance of real-world objects can be approximated utilizing accurate 3D models and modern rendering techniques such as in~\cite{Su.2015}. Here, the authors overlaid background images with object renderings, similar to the cut-and-paste approach but completely removing the necessity of annotated real-world data. 

Since real images are highly affected by the environment in which they are taken, the appearance of objects changes with their surroundings, impeding model prediction. Hence, a common approach is to randomize the environment of the objects of interest during image synthesis to train the DL model to deal with these variations. However, the authors of~\cite{Tobin.2017} hypothesized that complete DR, including even unrealistic variations of the objects of interest, helps to close the sim-to-real domain gap as the DL model is forced to focus on the object's shape rather than its color or glossiness. Therefore, they also randomized the textures of the 3D models and obtained promising results in the localization of simple geometric objects, enabling a robot to grab these objects in a real-world test~\cite{Tobin.2017}.

Based on this idea, several experiments have been conducted, assessing different degrees of DR and how domain knowledge can be used during simulation to improve real-world performance. For instance,~\cite{Prakash.2019} introduced context splines to the synthetic scenes to ensure that cars are always positioned on streets and~\cite{Mayershofer.2021} proposed to define object relations via configuration files to set the objects of interest into more realistic contexts. However, these methods were evaluated on small individual test cases, so the results are hardly comparable and also partly contradicting. This was addressed by the authors of~\cite{Eversberg.2021} who compared numerous DR approaches in the application of detecting turbine blades at manual working stations to provide a guideline on designing image synthesis pipelines for industrial use cases. They observed it to be most beneficial to focus on realistic lighting conditions and object textures while the image background and distractors can be randomized completely~\cite{Eversberg.2021}.

Complementarily, in our research, we follow these guidelines to develop a general strategy for pipeline design in complex industrial parts detection scenarios that ensures a manageable implementation effort and demonstrate the sim-to-real performance of state-of-the-art object detection models in the challenging industrial application of terminal strip object detection, representatively for these scenarios.

\section{Image Synthesis Pipeline}
\label{sec:pipeline}
\begin{figure}[!ht]
    \centering
    \includegraphics[width=0.95\linewidth]{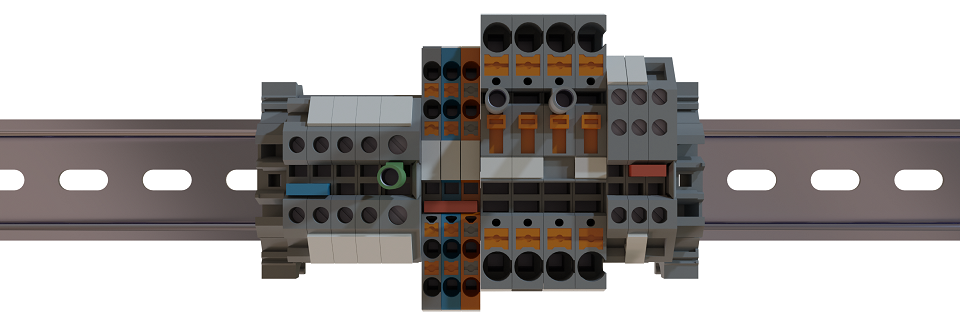}
    \caption{Example rendering of a terminal strip randomly generated by the process described in Subsection~\ref{subsec:creation}. It shows the centered front view that is most informative in this use case.}
    \label{fig:ts_example}
\end{figure}

The image synthesis pipeline takes 3D models and outputs annotated synthetic images of terminal strips, exploiting domain knowledge to set the objects of interest into a more realistic context. In this application, the most general characteristics we focus on are the following:

\begin{itemize}
    \item Terminal strips consist of several terminal blocks mounted next to each other on a DIN rail.
    \item The terminal blocks are equipped with additional accessories such as plug-in bridges, test adapters, markings and end covers.
    \item Neighboring terminal blocks often are of the same type, resulting in a group-like structure.
    \item Terminal strips are mounted on flat surfaces such as back panels of electrical cabinets.
    \item The most informative and usually the only accessible view of a terminal strip is the front view, as illustrated in Figure~\ref{fig:ts_example}.
\end{itemize}

Since it has been shown to be beneficial to include domain knowledge in DR~\cite{Prakash.2019,Mayershofer.2021}, we use this information in our pipeline along with standard randomization techniques to obtain random synthetic images that are contextually close to reality. In this section, we provide details about the different generation steps we implemented using the open-source rendering software Blender\footnote{Blender 3.3: \url{https://www.blender.org}}.

\subsection{Creation of Terminal Strips}
\label{subsec:creation}
In this paper, we consider 36 terminal blocks, including two connection technologies, three wire sizes, and two to four connection points to cover almost identical and easily distinguishable objects. Furthermore, the terminal blocks are randomly equipped with plug-in bridges, test adapters, markings, and end covers, and different realistic textures are used. Therefore, we manually created virtual materials by adapting roughness, specular, metallic, and RGB values in Blender until the renderings looked similar to real reference images.

The first step for creating a terminal strip regarding domain knowledge is to place the 3D model of a DIN rail in the center of the virtual scene. After that, uniformly sampled terminal blocks are virtually mounted next to each other on the DIN rail until the number of objects exceeds a threshold that restricts the size of the terminal strip. Each selected 3D model of a terminal block is thereby inserted up to five times to model the group-like structure of terminal strips, and the created virtual materials are randomly assigned. Since gray housings are more common in reality than those of other colors, we set the probability for the gray virtual material to 80\%, while the remaining colors uniformly share 20\%. Moreover, most real-world terminal blocks are one-sided open, requiring an end cover if the neighboring terminal block does not entirely cover the open side to ensure electrical insulation. This is implemented in the pipeline by inserting an end cover every time the height or depth of the selected terminal block is less than 90\% of the previous one. 

Apart from wiring, the main distractors of terminal strips are their additional accessories, such as plug-in bridges, test adapters, and markings. Although the authors of~\cite{Eversberg.2021} found that realistic distractors do not necessarily improve the performance of DL models, we decided to stay with realism for two reasons. First, the modeling effort is almost identical to simple, unrealistic distractors, as the required 3D models are available, and the virtual materials are similar to the materials used for the terminal blocks. Second, some applications may focus on detecting these accessories, e.g., to determine whether plug-in bridges are plugged correctly. Thus, each terminal strip is randomly equipped with markings, plug-in bridges, and test adapters according to its configuration.

Starting with the markings, we decide for every marking point individually whether the 3D model of the corresponding marking is inserted by sampling from a Bernoulli distribution with a probability of success of 70\%. In contrast, randomly placing plug-in bridges is more difficult since not all types of terminal blocks can be bridged, and the length of plug-in bridges is variable. Hence, we iterate over all shafts and determine the bridgeable ones to sample realistic bridge positions and different lengths afterward. Furthermore, red and blue plug-in bridges are considered with a probability of 70\% for red, as this color appears more often in reality. Regarding test adapters, the 3D models are randomly placed in the remaining shafts with a probability of 10\%, varying between four colors, and 20\% of the terminal strips are rendered without test adapters. 

Finally, terminal strips are often enclosed by so-called end clamps, so the pipeline is implemented to generate 50\% of the terminal strips containing a starting and an ending 3D model of an end clamp. Figure~\ref{fig:ts_example} shows a sample rendering of a terminal strip generated by the described randomization procedure.

\subsection{Viewpoints}
Considering different viewpoints in object detection and related ML-based CV tasks is essential, as the objects of interest usually appear arbitrarily rotated and scaled in real-world images. However, domain knowledge often enables the exclusion of unrealistic settings during image synthesis, reducing the number of images required to cover frequently occurring viewpoints. As mentioned previously, the most informative view of terminal strips is their front view, illustrated in Figure~\ref{fig:ts_example}, which also is the only available view when the terminal strip is already mounted on a flat surface. Although in a production environment, such as in quality control, a camera could be fixed in a well-centered position, slightly varying viewpoints occur when an image is taken by hand, e.g., in front of an electrical cabinet, which is the more general case we focus on. 

\begin{figure}[!ht]
    \centering
    \includegraphics[width=0.95\linewidth]{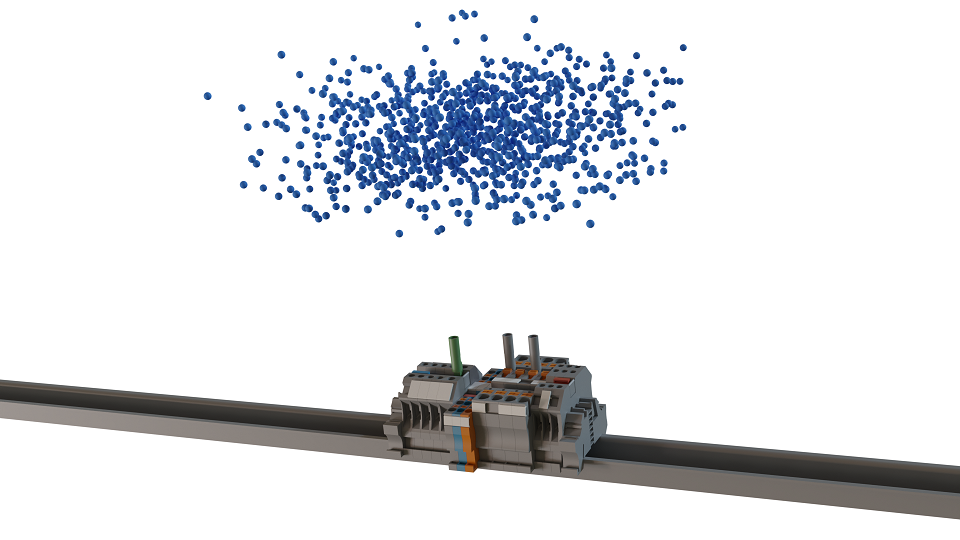}
    \caption{Visualization of 1000 sampled camera positions in the form of a point cloud to illustrate the procedure of randomizing viewpoints.}
    \label{fig:sampled_cam_pos}
\end{figure}

In our pipeline, we place a virtual camera in Blender perfectly centered in front of the generated terminal strip and directed to its center to start with the optimal front view. Then, we independently add 3D Gaussian noise to the camera's position and orientation, resulting in a majority of viewpoints close to the reference view while reasonably deviating viewpoints are also covered. As a side effect, this approach captures slightly different scales of the objects of interest, as the distance between the camera and the terminal strips varies. To show the outcome of this viewpoint randomization, Figure~\ref{fig:sampled_cam_pos} visualizes 1,000 sampled camera positions in the form of a point cloud.

\subsection{Lighting Conditions}
In recent image synthesis methods for visual DL applications, there are two common approaches to randomize the lighting conditions: using parameterizable, virtual light sources and implementing image-based lighting. The former is very flexible and allows accurate modeling of the desired illumination. However, creating real-world conditions is laborious since light rays of different colors and intensities might enter the scene from various directions, and reflections should also be considered. In contrast, image-based lighting maps a $360^{\circ}$ image on a globe and uses the pixel values to illuminate the 3D models as if they were in the captured environment. This is less customizable but reduces the creation of complex lighting conditions to the selection of suitable images and has shown to be superior when only little effort is invested in parameterizing virtual light sources~\cite{Eversberg.2021}. 

\begin{figure}[!ht]
    \centering
    \includegraphics[width=0.95\linewidth]{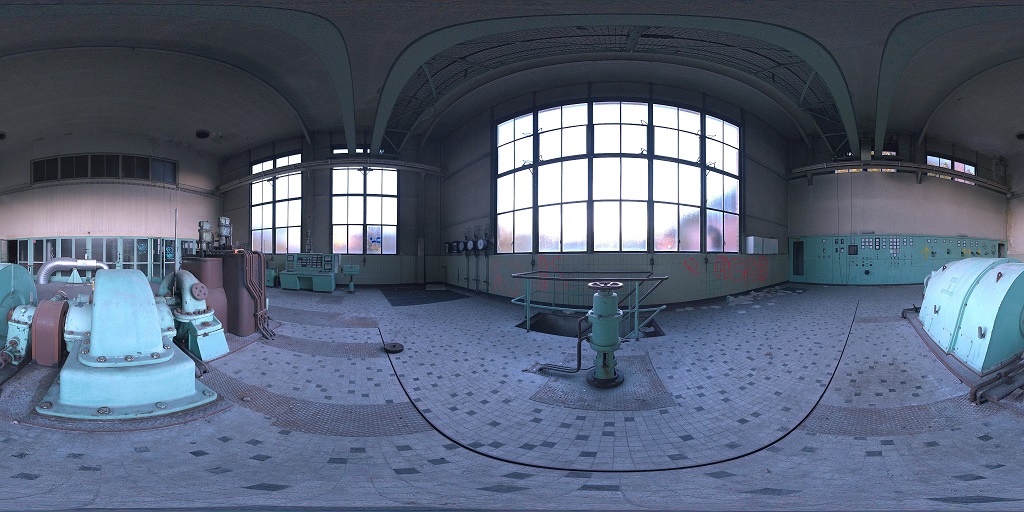}
    \caption{Example $360^{\circ}$ HDRI from Poly Haven used for the image-based lighting.}
    \label{fig:example_hdri}
\end{figure}

To keep the modeling effort small, we selected 46 high dynamic range images (HDRIs) from Poly Haven\footnote{\url{https://polyhaven.com/hdris}} representing $360^{\circ}$ indoor scenes with an industrial context or medium contrast, as shown in Figure~\ref{fig:example_hdri}. During dataset creation, a new HDRI is uniformly sampled for each generated terminal strip and imported such that the back side of the DIN rail faces the ground of the captured environment. This is reasonable since it excludes the presence of light sources behind the DIN rail corresponding to the real-world conditions when the terminal strip is mounted, e.g., in an electrical cabinet. To further increase the variety of possible illuminations, the HDRIs are randomly rotated around the vertical axis with a resolution of $1^{\circ}$ so that $46\cdot360=16,560$ at least slightly different lighting conditions can occur in the resulting dataset.

\subsection{Image Background}
Similar to the lighting conditions, synthetic image backgrounds can be virtually created and rendered along with the objects of interest~\cite{Akar.2022} or they rely on real images~\cite{Eversberg.2021}. On the one hand, modeling the entire background leads to more realistic images, as all objects can be related, and visualizing realistic shadows is possible. On the other hand, random image-based backgrounds have shown promising results~\cite{Eversberg.2021} and are less complex, which is more suitable for industrial manufacturers. Nevertheless, the authors of~\cite{Alghonaim.2021} found that realistic shadows significantly improve the performance of DL models in a 6D pose estimation task. Therefore, we decided to follow a compromising strategy, reusing the HDRIs of the environment lighting and visualizing the shadow of the terminal strip as if it were mounted on a flat surface.

\begin{figure}[!ht]
    \centering
    \includegraphics[width=0.95\linewidth]{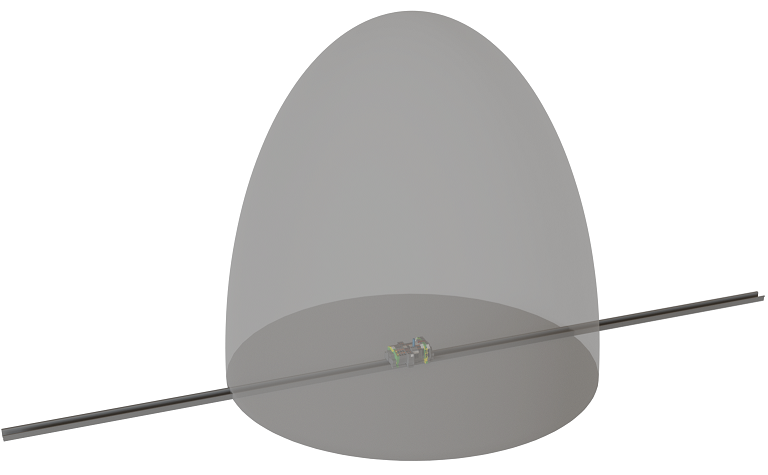}
    \caption{Illustration of the semi-ellipsoid used to visualize the shadows of the terminal strips and onto which the HDRIs are mapped to create the image background.}
    \label{fig:shadow_catcher}
\end{figure}

The implementation is based on a surrounding semi-ellipsoid, as illustrated in Figure~\ref{fig:shadow_catcher}, that serves as a shadow catcher. It is positioned so that the flat side represents the mounting surface for the DIN rail to which the terminal strip is centered, allowing the objects of interest to cast shadows on the background. To insert the HDRI, the same image from the environment lighting is mapped onto the surface of the semi-ellipsoid, and the respective rotation around the vertical axis is applied. As a result, the background of a synthetic image consists of the ground of the environment captured by the HDRI with additional shadows cast by the terminal strip, which is more realistic than using fully randomized backgrounds and less expensive than modeling the entire scene. See Appendix~\ref{sec:A} for synthetic sample images generated using this approach.

\subsection{Annotations}
\begin{figure}[!ht]
    \centering
    \includegraphics{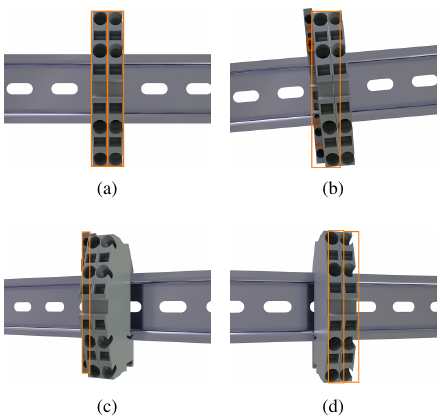}
    \caption{Four different annotation scenarios that illustrate the difficulty of determining accurate bounding boxes for narrow objects, even in image synthesis.} 
    \label{fig:BBScenarios}
\end{figure}

The main advantage of generating synthetic images for supervised learning is that ground-truth annotations can be calculated automatically and almost for free. In object detection tasks, these annotations consist of a class label and the position of the object, represented as an axis-aligned bounding box. However, even in synthetic environments, accurate bounding boxes can be hard to provide in so-called crowded scenes, meaning that objects are densely arranged and occlude each other. This also applies to terminal strips, as can be seen in Figure~\ref{fig:BBScenarios}. 

While scenario (a) shows an unproblematic case of a well-centered front view of two terminal blocks, image (b) demonstrates that a slight camera rotation of $5^{\circ}$ around the viewing direction already widens the bounding boxes, such that they include half of the neighboring terminal blocks. Moreover, scenarios (c) and (d) illustrate that components can occlude each other, and terminal blocks with a roof-like shape are hard to annotate, depending on the camera position.

The comparison shows that including all visible parts of a terminal block into its bounding box inevitably leads to highly overlapping annotations, which is critical for most DL-based detection models. The major problem arises from the non-maximum suppression (NMS) algorithm commonly used in object detection. It is a post-processing step that iteratively selects the detection with the highest prediction score and removes all highly overlapping detections with the same predicted class label based on a threshold. In crowded scenes, NMS often results in a large number of false negatives, i.e., undetected objects, which compromises the overall detection performance~\cite{Jiang.2023}. Therefore, we decided to restrict the overlaps of terminal block bounding boxes to 20\% of the width of the thinner component, encouraging the DL models to learn tight positional predictions with reasonable overlaps during training. Additionally, we provide annotations for the end clamps, end covers, plug-in bridges, and test adapters to enable applications in which accessories should be detected.

\section{Evaluation}
\label{sec:evaluation}
In this section, we evaluate the sim-to-real performance of standard object detection models in the industrial application of terminal strip object detection using the proposed image synthesis pipeline. Therefore, we selected the two standard model architectures, RetinaNet~\cite{Lin.2017} and Faster R-CNN~\cite{Ren.2015}, representatively for the common single-stage and two-stage detectors, and the two more recent models YOLOv8~\cite{yolov8_ultralytics} and DINO~\cite{zhang2022dino} to demonstrate robustness across different model architectures. For RetinaNet, Faster R-CNN, and DINO we used a ResNet50~\cite{He.2016} backbone.

The evaluation is based on the mean average precision (mAP) metrics mAP@0.5 and mAP@[0.5:0.95], as defined for the MS COCO~\cite{Lin.2014} challenge. While the first considers a single intersection over union (IoU) threshold of 0.5, the second averages over ten equidistant IoU thresholds ranging from 0.5 to 0.95 to evaluate the predicted position of the objects of interest more strictly. Thus, it can be distinguished between an overall poor detection performance and correct detections with slightly inaccurate bounding boxes. We abbreviate mAP@[0.5:0.95] to mAP in the following for readability.

\subsection{Datasets}
For our experiments, we generated a synthetic dataset of 30,000 randomly configured terminal strip images at a resolution of $1024\times512$ pixels using the proposed image synthesis pipeline. It contains 36 frequently occurring terminal blocks in five different colors and suitable plug-in bridges, test adapters, end covers, and end clamps as additional accessories. The markings are only considered as distractors and not part of the detection, so 40 detection classes, composed of very similar and well-separable objects, are provided in total. The average number of objects per image is 25, whereas the accessories are overrepresented compared to the evenly distributed terminal blocks, as they occur in almost every image. The dataset is split into 79\% train, 1\% evaluation, and 20\% test images to enable testing on synthetic images required to compare the performances of the DL models in simulation and reality. Although a 1\% evaluation split may appear small and unconventional, we found it to be sufficient as it still covers around 7,500 detection objects. Moreover, we did not intend to excessively tune hyperparameters.

To perform the real-world test, we used the first part of the image synthesis pipeline, as described in Subsection~\ref{subsec:creation}, to obtain 100 parts lists of random terminal strips and build them in reality. After that, we took three pictures of each terminal strip from different viewpoints and manually labeled the resulting 300 images for object detection. Hence, the image contents in the synthetic and the real-world datasets are very similar, enabling a meaningful investigation of the sim-to-real domain gap. This dataset is far more complex than the datasets in previous studies of the sim-to-real generalization performance and gives reasonable insights into the potential of synthetic training images in real-world industrial applications. 

\subsection{Experiments}
In our experiments, we leveraged the model implementations from the Open MMLab Detection Toolbox~\cite{mmdetection} In this framework, the models can be configured via configuration files and several checkpoints, e.g., from MS COCO pre-training, are available. Since we aim to provide the performance of standard DL models, we kept the default configurations except from the following changes:

\begin{itemize}
    \item The target size of the preprocessing image resizer was adapted to the size of the synthetic images.
    \item All on-the-fly data augmentation methods were removed since the dataset could be extended by generating more synthetic images if necessary, and further variations are also realizable in the image synthesis pipeline.
    \item If present, the aspect ratios of the anchor boxes were adapted to the most frequent bounding box ratios in the synthetic training dataset.
    \item MS COCO pre-training was selected, even though we obtained similar results with ImageNet pre-training.
\end{itemize}
\noindent
Using these configurations, the selected models were trained for 20 epochs with a batch size of 16 on a single NVIDIA A100 GPU with 80 GB of memory. To quantify the sim-to-real domain gap, we then evaluated the trained models on the synthetic test set and the manually labeled real images. 

As we found the scale of the real images to have a high impact on the model performance, we further experimented with rescaling the real-world data. This sensitivity stems from the fact that standard CNNs are not equivariant to the dilatation group ($\mathbb{R}^+$, $\times$), meaning their performance degrades significantly under scale variations not seen during training. Therefore, we considered three different approaches to determine the rescaling factors for preprocessing. First, a constant factor of 1.5 was applied to all real images, as it seemed to scale the majority into the range of the synthetic images. Second, the image synthesis pipeline was used to generate 1,000 additional terminal strip images of various scales annotated with the corresponding rescaling factors to fine-tune a simple ResNet50 regression model from ImageNet pre-training and predict the scaling factor for each real image independently. Finally, we also explored the detection performance when scaling factors are irrelevant, such as in settings with a fixed camera position. To address this, we further employed Bayesian optimization with Gaussian Processes to automatically determine the optimal scaling factor for each real test image individually. In this formulation, the scaling factor was treated as a continuous optimization variable, and the objective function was defined as the F1-score of the detector on the given image. This approach serves two purposes: first, it provides an upper bound on the sim-to-real generalization performance when the effects of scale mismatch are mitigated, and second, it allows us to evaluate the ResNet50 regression model. We refer to these experiments as real (1.5), real (ResNet), and real (Opt) in the following.

\subsection{Results}
The performances of RetinaNet, Faster R-CNN, YOLOv8, and DINO in our experiments are listed in Table \ref{tab:Results}. All models reach 99\% $mAP@0.5$ and more than 94\% $mAP$ on the synthetic test images, showing that terminal strip object detection is well-solvable in the simulated domain, and even very similar terminal blocks can be distinguished. Considering this as the reference performance, the sim-to-real domain gap seems to be huge in the case of unscaled real test images since the best DINO model only achieves 87.30\% $mAP@0.5$, and the performance drop is even more significant for the other models.

\begin{table*}[t]
    \centering
    \caption{Comparison of the sim-to-real detection performances of RetinaNet, Faster R-CNN, YOLOv8, and DINO based on different scaling methods for image preprocessing. The inference times were measured as the mean of 100 inference runs on random images of the real test dataset.}
    \renewcommand{\arraystretch}{1.3}
    \resizebox{\textwidth}{!}{
    \begin{tabular}{l|cc|cc|cc|cc}
        & \multicolumn{2}{c|}{\textbf{RetinaNet}} & \multicolumn{2}{c|}{\textbf{Faster R-CNN}} & \multicolumn{2}{c|}{\textbf{YOLOv8}} & \multicolumn{2}{c}{\textbf{DINO}}\\
        Test dataset & $mAP$ & $mAP@0.5$ & $mAP$ & $mAP@0.5$ & $mAP$ & $mAP@0.5$ & $mAP$ & $mAP@0.5$\\
        \hline
        synthetic       & $94.30\%$ & $99.60\%$ & $96.60\%$ & $99.40\%$ & $97.30\%$ & $99.40\%$ & $97.80\%$ & $99.90\%$\\
        \hdashline
        real            & $43.60\%$ & $65.00\%$ & $50.70\%$ & $76.90\%$ & $29.60\%$ & $43.70\%$ & $57.60\%$ & $87.30\%$\\
        real (1.5)      & $58.60\%$ & $85.80\%$ & $64.80\%$ & $95.30\%$ & $58.20\%$ & $84.80\%$ & $65.50\%$ & $95.60\%$\\
        real (ResNet)   & $63.60\%$ & $93.00\%$ & $64.60\%$ & $96.00\%$ & $63.00\%$ & $92.40\%$ & $65.90\%$ & $97.40\%$\\
        real (Opt)      & $66.70\%$ & $97.70\%$ & $66.30\%$ & $98.30\%$ & $66.80\%$ & $97.80\%$ & $66.70\%$ & $98.40\%$\\
        \hline
        inference time & \multicolumn{2}{c|}{$23.24 ms$} & \multicolumn{2}{c|}{$24.81 ms$} & \multicolumn{2}{c|}{$23.83 ms$} & \multicolumn{2}{c}{$43.73 ms$}
    \end{tabular}
    \label{tab:Results}
    }
    \renewcommand{\arraystretch}{1}
\end{table*}

However, we find many terminal blocks to be misclassified as their smaller-sized variant, e.g., PT~4 as PT~2,5 or PT~4-TWIN as PT~2,5-TWIN (see Appendix~\ref{sec:A}). Since this observation does not correspond to the results of the synthetic test and the real-world dataset tends to show the terminal strips on a smaller scale than the synthetic dataset, we assume that the size of the terminal strip components in the image is crucial to the distinction of similar objects that mainly differ in size. Hence, we briefly test this hypothesis by resizing all real images with a constant scaling factor of 1.5 and obtain the expected improvement with all models exceeding 84\% $mAP@0.5$ and DINO even reaching 95.60\%. While in other use cases the occurrence of different object scales could be addressed by varying object sizes during training data generation, this is no option when some object classes only differ in size. This necessitates an approach with only small scale variations in training and a dynamic scaling strategy during inference. 

One possibility for dynamically determining the scaling factor for a terminal strip image is to use a preprocessing DL model to predict it. Therefore, we again rendered synthetic images to train a ResNet50 regression model, accepting the additional influence of the domain gap on the detection performance. Utilizing this model to resize the real test images further improves the detection performances of all considered detectors with Faster R-CNN and DINO reaching 96.00\% and 97.40\% $mAP@0.5$ respectively. This quantifies the domain gap to 3.40\% and 2.50\% for these models at this point , which is a remarkable result.

However, the dependency of the performance on the object scale is very application-specific and does not transfer to any industrial parts detection application. To this end, we applied Bayesian optimization with Gaussian Processes to each model and each real test image individually to approach the detection performance when scaling is aligned in both domains. The results can be seen in the penultimate row of Table~\ref{tab:Results}, where all models reveal a sim-to-real domain gap of less than 2\% in $mAP@0.5$. This demonstrates the power of combining DR with domain knowledge even in the challenging industrial application of terminal strip object detection.

\begin{figure}[!ht]
    \centering
    \includegraphics[width=0.95\linewidth]{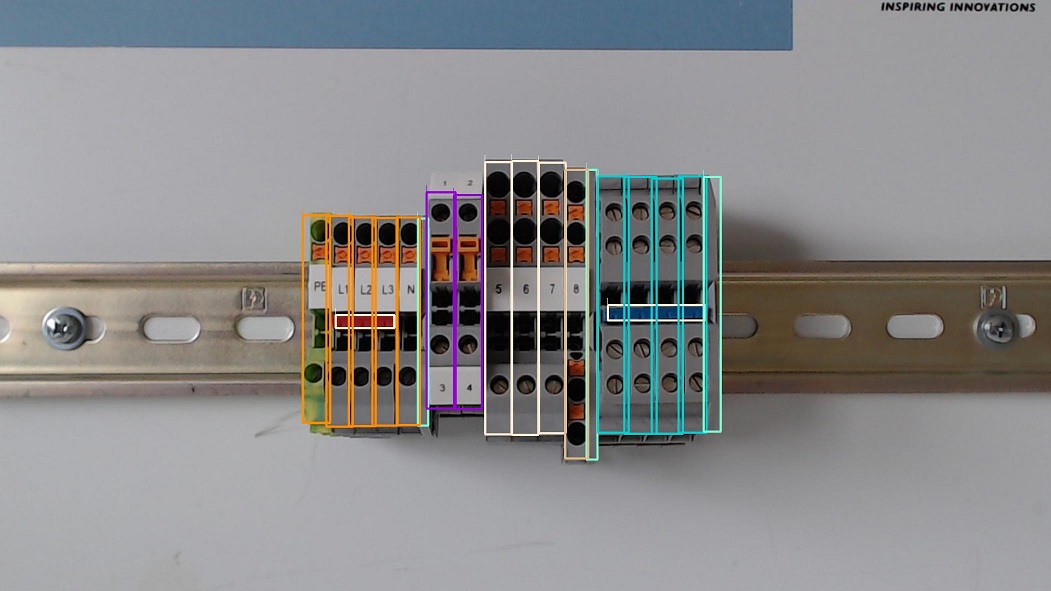}
    \caption{Example prediction of Faster R-CNN under optimized scaling conditions showing inaccurate bounding boxes caused by the low contrast between markings and background.}
    \label{fig:inaccurate_bbs}
\end{figure}

The main performance differences between synthetic and real test images seem to lie in the accuracy of the predicted bounding boxes, as the mAP score does not exceed 66.80\% in any experiment. Looking at the predictions, we observe that the white background often causes inaccurate bounding boxes when a terminal block is equipped with a marking at its upper or lower edge so that there is a low contrast between the object and background, as depicted in Figure~\ref{fig:inaccurate_bbs}. Furthermore, we belief that the manual annotations are not as exact as the calculated ones, introducing additional inaccuracies. Hence, the lower performance in mAP is explainable and could be addressed in real-world applications, e.g., by ensuring backgrounds deviating in color.

\section{Discussion}
\label{sec:discussion}
Our studies show that real-world industrial object detection tasks come with their own characteristics and challenges that need to be addressed to introduce recent advances in DL to industrial environments. In particular, industrial settings often involve visually similar objects, repetitive arrangements, special object shapes, limited annotated data, and strict requirements for reliability and scalability. While previous research was more focused on the effect of individual image synthesis and randomization methods in simple domains rather than providing a usable pipeline for complex industrial applications, we developed a strategy that ensures a manageable implementation effort and demonstrated its strong baseline performance on this kind of applications.

However, there are a few things to consider to transfer our approach to a different industrial parts detection application. Assuming that detailed 3D models and expert knowledge about typical arrangements are available, it can be challenging to create realistic virtual materials that not only look like the real industrial parts in terms of color, but also have the same properties, e.g. reflection. This took us a few iterations as well, but one could also define a range of values and randomize the material during dataset generation to ensure robustness. Furthermore, the environment should be taken into account. In a more general case where the objects of interest can occur under various lighting conditions and in front of diverse backgrounds, it may be favorable to use a lot of different backgrounds and light settings in the training data, whereas under laboratory conditions the setups in reality and simulation can be aligned more closely to improve the performance.

In general, these advices together with the realistic arrangement of objects and the environment randomization already enable a cost effective generation of a synthetic image dataset for industrial parts detection. Nevertheless, application-specific issues can occur as we faced with scaling and object classes that only differ in size. The results show that these issues can harm the detection performance substantially. However, this gave us the chance to demonstrate that the same strategy can also be applied, e.g., to learn a suitable image preprocessing, even if our approach for predicting scaling factors leaves room for improvement. Regardless, we expect that this strategy can also help to overcome other application-specific issues without major changes.

As in limited data scenarios, data augmentation is another common way of extending the training dataset, the pros and cons of rendered data vs. augmented data should also be briefly discussed. While both approaches create synthetic images in a sense, a key difference is that data augmentation requires a base dataset of labeled images. On condition that no ground truth data of the target domain is available, which we focused on as a worst case consideration, this base dataset could only consist of synthetic images. Under this premise, data augmentation techniques such as variations in color, contrast or brightness could also be included in the simulation directly by changing the respective parameters in the rendering software. Hence, the only advantage of data augmentation would be that it is usually faster than rendering which could be relevant on a large scale. Regardless, augmentations must be selected as carefully as the randomization in the image synthesis pipeline to reflect domain-specific constraints. E.g. it would not make sense to randomly flip the images, as in our application terminal strips only occur in a single orientation. To support our assumptions on this, we performed an additional experiment in which we replaced $\frac{1}{3}$ of the original training data with some reasonable augmentations. The details and results can be found in Table~\ref{tab:ResultsAug} in the appendix.

It is important to note that we intend to provide a lower bound of the expectable detection performance in these complex industrial applications, which still can be improved. For instance, any ground truth data of the target domain is expected to be superior to synthetic data and to potentially further improve the performance. On that occasion, it would be interesting to benchmark augmented data of the target domain against synthetic data, which we leave for future work. To conclude, we think our results will serve as evidence for industrial manufacturers to start proof-of-concepts for their own use cases, as the cost-benefit ratio is promising.

\section{Conclusions}
\label{sec:conclusions}
In this paper, we investigated the impact of synthetic training data in the challenging industrial application of terminal strip object detection, which is representative for many detection tasks of industrial manufacturers. These tasks are often characterized by similar objects, repetitive arrangements, special object shapes, limited annotated data, and strict requirements for reliability and scalability discouraging companies from adopting DL solutions. By leveraging common randomization techniques and carefully incorporating domain knowledge into the image synthesis process, we demonstrated that it is possible to reach a detection performance suitable for industrial requirements with a manageable implementation effort. We also show that our approach is robust across model architectures, whereas the transformer based DINO model showed the best performance with 98.40\% $mAP@0.5$ on our real-world test set.

We discussed the domain-specific issue of various scales in combination with objects that only differ in size, where we demonstrated that our strategy can be applied to learn a suitable image preprocessing from synthetic training data. This indicates that our approach is adaptable to various tasks in this context, although our learned image rescaling could not fix the issue entirely. We will further investigate this in future work. Moreover, additional object classes should be considered to scale up our approach, as product portfolios of industrial manufacturers often consist of more than 40 components, such as Phoenix Contact offering around 10,000 parts related to terminal strips. Finally, we plan to investigate orthographic images of terminal strips since perspective images complicate the determination of accurate bounding boxes in the edge regions.

In summary, our work establishes a practical and extensible strategy for leveraging fully synthetic data in complex industrial parts detection applications. It encourages manufacturers to consider image synthesis pipelines as a viable alternative to costly manual data collection and annotation, and provides a foundation for future advances in sim-to-real learning for industrial automation. In comparison to data augmentation, our approach has proven to be a strong alternative when no ground truth data of the target domain is available. Furthermore, we made our dataset publicly available for reference and future research to support further developments in this area.

\backmatter

\bibliography{sn-bibliography}
\newpage
\onecolumn
\begin{appendices}

\section{}
\label{sec:A}

\begin{figure}[!ht]
    \centering
    \begin{minipage}[t][0.38\textheight][t]{0.49\textwidth}
        \centering
        \includegraphics[height=0.35\textheight]{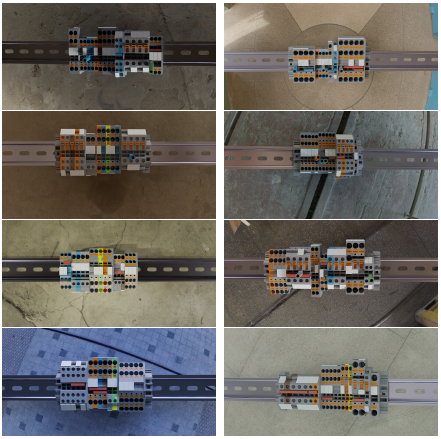}
        \caption{Randomly selected synthetic training images generated by the proposed image synthesis pipeline.}
        \label{fig:synthetic_images}
    \end{minipage}\hfill
    \begin{minipage}[t][0.38\textheight][t]{0.49\textwidth}
        \centering
        \includegraphics[height=0.35\textheight]{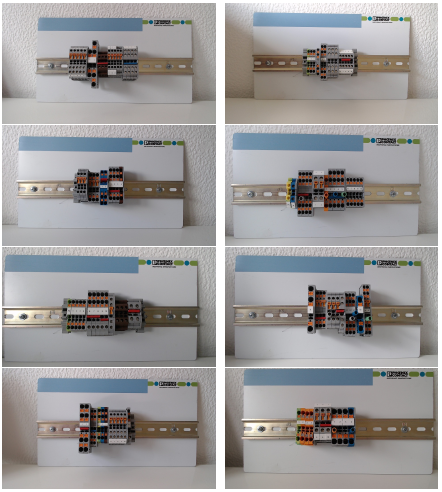}
        \caption{Randomly selected images from the real-world dataset.}
        \label{fig:real_images}
    \end{minipage}
\end{figure}

\begin{table*}[h!]
    \centering
    \caption{Comparison of the sim-to-real detection performances of RetinaNet, Faster R-CNN, YOLOv8, and DINO based on different scaling methods for image preprocessing. For this experiment $\frac{1}{3}$ of the training data was replaced by data augmentation to investigate the impact of more generated data vs. augmented data. For dataset augmentation we used random brightness and contrast variations, Gaussian noise, motion blur, image compression, and slight perspective variations.}
    \renewcommand{\arraystretch}{1.3}
    \resizebox{\textwidth}{!}{
    \begin{tabular}{l|cc|cc|cc|cc}
        & \multicolumn{2}{c|}{\textbf{RetinaNet}} & \multicolumn{2}{c|}{\textbf{Faster R-CNN}} & \multicolumn{2}{c|}{\textbf{YOLOv8}} & \multicolumn{2}{c}{\textbf{DINO}}\\
        Test dataset & $mAP$ & $mAP@0.5$ & $mAP$ & $mAP@0.5$ & $mAP$ & $mAP@0.5$ & $mAP$ & $mAP@0.5$\\
        \hline
        synthetic       & $93.30\%$ & $98.80\%$ & $96.10\%$ & $99.30\%$ & $96.70\%$ & $99.30\%$ & $97.40\%$ & $99.90\%$\\
        \hdashline
        real            & $39.10\%$ & $58.70\%$ & $47.80\%$ & $71.90\%$ & $22.80\%$ & $34.80\%$ & $55.60\%$ & $86.20\%$\\
        real (1.5)      & $59.00\%$ & $86.30\%$ & $64.80\%$ & $94.70\%$ & $56.90\%$ & $86.70\%$ & $65.50\%$ & $98.10\%$\\
        real (ResNet)   & $65.10\%$ & $93.60\%$ & $66.10\%$ & $96.40\%$ & $61.60\%$ & $91.00\%$ & $64.80\%$ & $98.40\%$\\
        real (Opt)      & $67.30\%$ & $96.60\%$ & $67.10\%$ & $98.00\%$ & $64.60\%$ & $96.50\%$ & $65.40\%$ & $98.70\%$
    \end{tabular}
    \label{tab:ResultsAug}
    }
    \renewcommand{\arraystretch}{1}
\end{table*}

\begin{figure}[!ht]
    \centering
    \includegraphics[width=\textwidth]{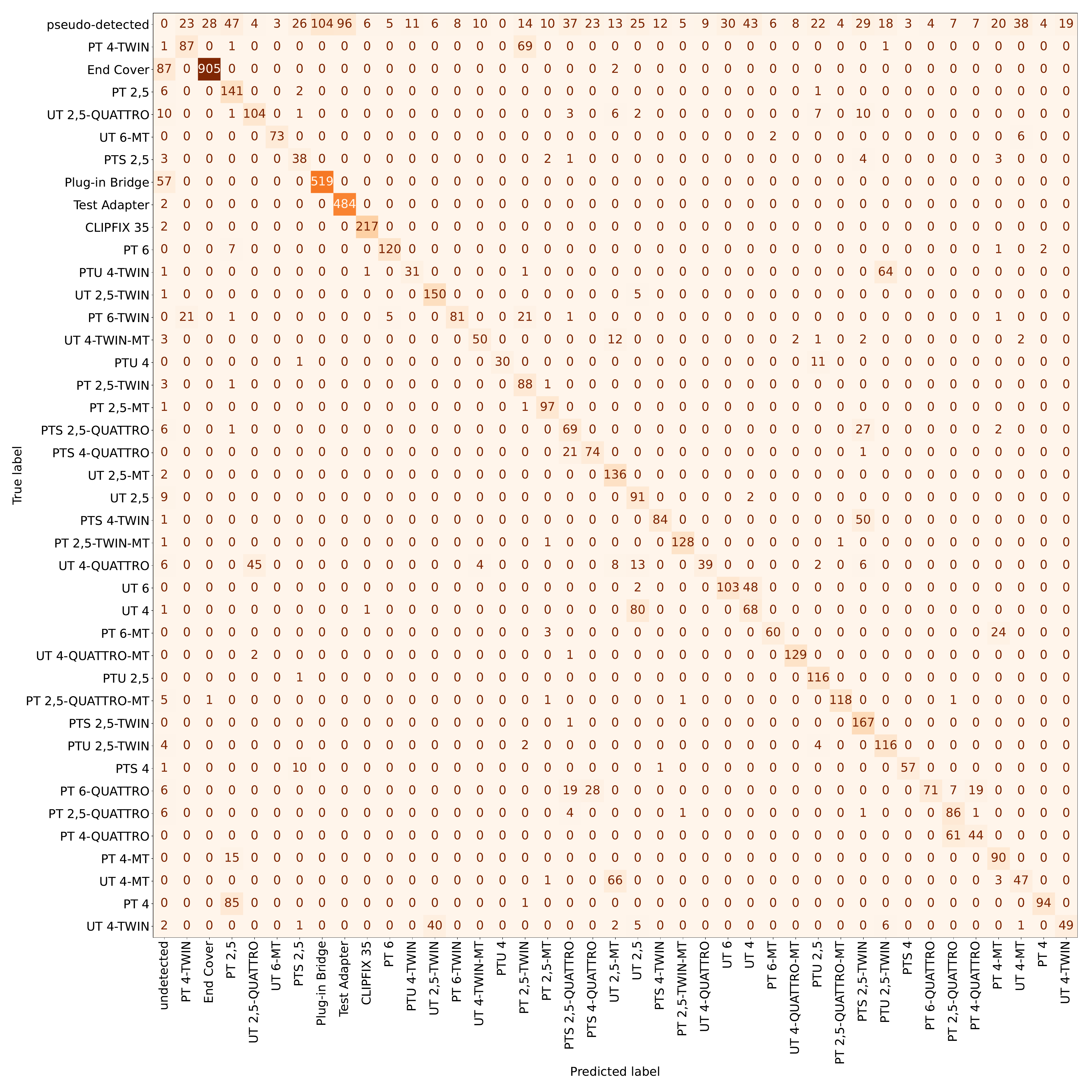}
    \caption{Confusion matrix of the Faster R-CNN detection model evaluated on the unscaled real images at IoU and score thresholds of 0.5. While many correct detections can be seen on the diagonal, certain misclassifications often occur. Among them are terminal blocks such as PT 4 or UT 6 that are repeatedly misclassified as their smaller-sized variants. Although we expected these types of false predictions, they could not be observed in the synthetic tests, indicating a sim-to-real discrepancy in scaling.}
\end{figure}




\end{appendices}

\end{document}